\documentclass[conference]{IEEEtran}
\IEEEoverridecommandlockouts
\usepackage{cite}
\usepackage{amsmath,amssymb,amsfonts}
\usepackage{algorithmic}
\usepackage{graphicx}
\usepackage{textcomp}
\usepackage{xcolor}
\usepackage{booktabs}
\usepackage{url}
\usepackage{CJKutf8}
\usepackage[colorlinks]{hyperref}
\usepackage{makecell}
\usepackage{multicol}

\newcommand{\tb}[1]{\textbf{#1}}

\def\BibTeX{{\rm B\kern-.05em{\sc i\kern-.025em b}\kern-.08em
    T\kern-.1667em\lower.7ex\hbox{E}\kern-.125emX}}
\begin{document}

\makeatletter
\newcommand{\linebreakand}{
  \end{@IEEEauthorhalign}
  \hfill\mbox{}\par
  \mbox{}\hfill\begin{@IEEEauthorhalign}
}
\makeatother

\title{Evaluating Large Language Models on Financial Report Summarization: An Empirical Study}

\author{
\IEEEauthorblockN{1\textsuperscript{st} Xinqi YANG}
\IEEEauthorblockA{\textit{IRUCA Research and Development 1} \\
\textit{IRUCA Co.,LTD}\\
Tokyo, Japan \\
yang@iruca.ai}
\and
\IEEEauthorblockN{2\textsuperscript{nd} Scott ZANG}
\IEEEauthorblockA{\textit{IRUCA Research and Development 1} \\
\textit{IRUCA Co.,LTD}\\
Tokyo, Japan \\
scott.zang@iruca.ai}
\and
\IEEEauthorblockN{3\textsuperscript{rd} Yong REN}
\IEEEauthorblockA{\textit{IRUCA Research and Development 1} \\
\textit{IRUCA Co.,LTD}\\
Tokyo, Japan \\
renyong@iruca.ai}

\linebreakand 

\IEEEauthorblockN{4\textsuperscript{th} Dingjie PENG}
\IEEEauthorblockA{\textit{Graduate School of Fund. Sc. and Eng.} \\
\textit{Waseda University}\\
Tokyo, Japan \\
kefipher9013@asagi.waseda.jp}
\and
\IEEEauthorblockN{5\textsuperscript{th} Zheng WEN}
\IEEEauthorblockA{\textit{Faculty of Science and Engineering} \\
\textit{Waseda University}\\
Tokyo, Japan \\
robinwen@aoni.waseda.jp}
}

\maketitle

\begin{abstract}
In recent years, Large Language Models (LLMs) have demonstrated remarkable versatility across various applications, including natural language understanding, domain-specific knowledge tasks, etc. However, applying LLMs to complex, high-stakes domains like finance requires rigorous evaluation to ensure reliability, accuracy, and compliance with industry standards. To address this need, we conduct a comprehensive and comparative study on three state-of-the-art LLMs, GLM-4, Mistral-NeMo, and LLaMA3.1, focusing on their effectiveness in generating automated financial reports. Our primary motivation is to explore how these models can be harnessed within finance, a field demanding precision, contextual relevance, and robustness against erroneous or misleading information. By examining each model’s capabilities, we aim to provide an insightful assessment of their strengths and limitations. Our paper offers benchmark for financial report analysis, encompassing proposed metrics such as ROUGE-1, BERT Score, and LLM Score. We introduce an innovative evaluation framework that integrates both quantitative metrics (\textit{e.g.}, precision, recall) and qualitative analyses (\textit{e.g.}, contextual fit, consistency) to provide a holistic view of each model's output quality. Additionally, we make our financial dataset publicly available, inviting researchers and practitioners to leverage, scrutinize, and enhance our findings through broader community engagement and collaborative improvement. Our dataset is available on huggingface\footnote{\url{https://huggingface.co/datasets/xinqiyang/tradingview_msn_financial_news_1k}}.
\end{abstract}


\section{Introduction}


 \begin{figure}[tbp] 
    \begin{center}
        \includegraphics[width=0.5\textwidth]{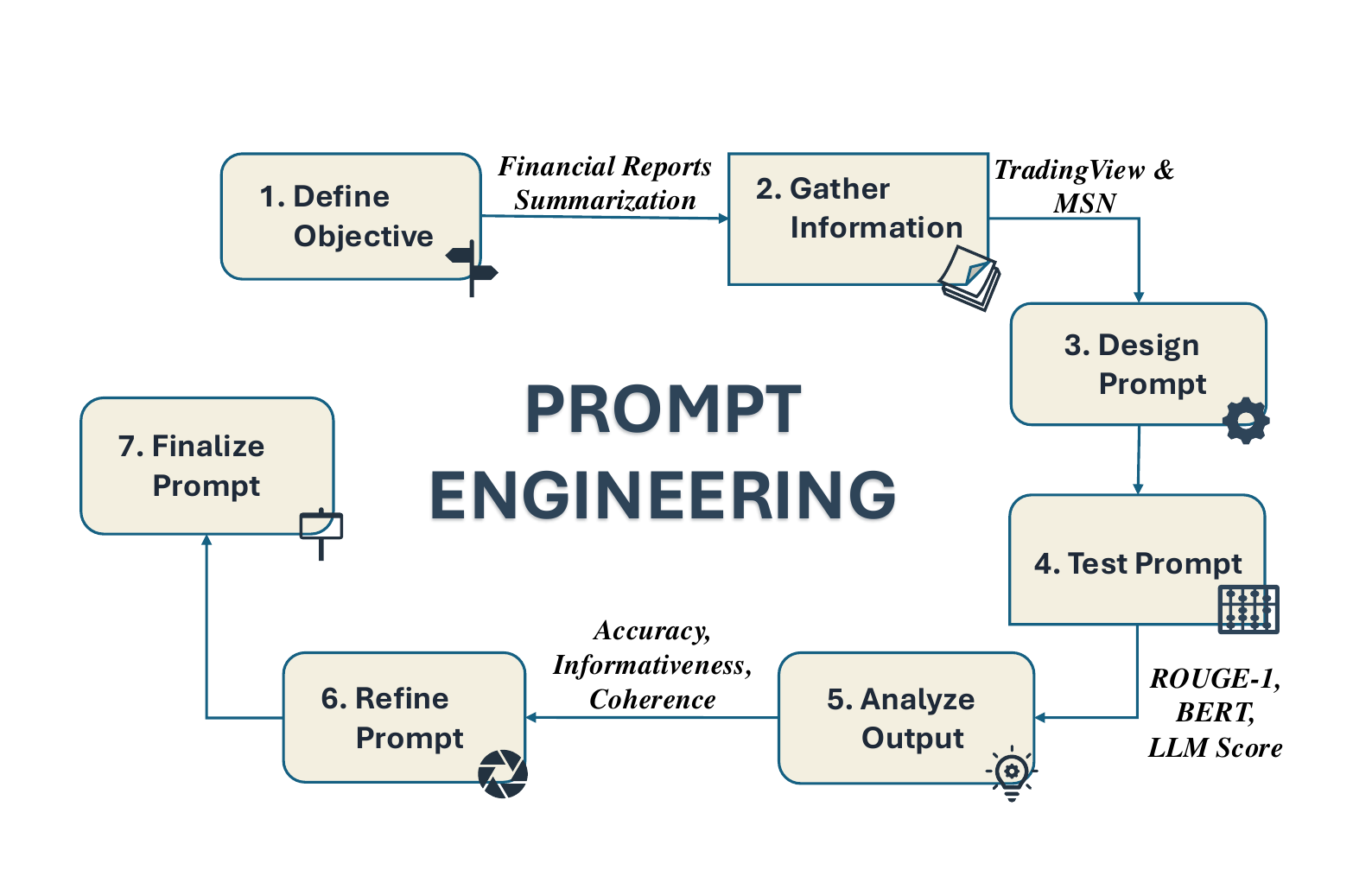}
    \end{center}
    \caption{The Flow Chart of Prompt Engineering in Financial Report Summarization.}
    \label{fig-PE}
\end{figure}

Large Language Models (LLMs)~\cite{trainllm, llama, gemma,phi3, TheC3,mamba,hpt} have emerged as transformative tools in Natural Language Processing (NLP), demonstrating unprecedented capabilities across tasks such as text generation~\cite{textgen}, summarization~\cite{textsum}, classification~\cite{textcls}, translation~\cite{llmtrans}, automatic dialogue~\cite{DialogueLLM}, and also extend to the multi-modal area~\cite{mulmodllm}. Built upon massive datasets and powered by advanced neural architectures, LLMs can understand and generate human-like text, making them adaptable to a wide range of domains. Recently, their application has extended into high-stakes sectors such as finance~\cite{finbert}, healthcare~\cite{health}, and legal industries~\cite{legal}, where \emph{accuracy}, \emph{informativeness} and \emph{coherence} are essential. However, deploying LLMs in such fields poses unique challenges, requiring careful evaluation to ensure reliability and trustworthiness.

The financial sector, in particular, benefits from LLM-driven innovations in automation, such as generating reports, analyzing market sentiment, and summarizing extensive financial data~\cite{fingpt}. Yet, financial text is often dense with specialized terminology and context-sensitive information, demanding high levels of precision and contextual understanding from LLMs. Consequently, assessing LLM performance in financial applications goes beyond conventional metrics, requiring a nuanced approach that includes \emph{accuracy}, \emph{informativeness} and \emph{coherence}, as well as alignment with financial domain knowledge. 

This paper presents a comprehensive evaluation of our strive of prompt engineering on three cutting-edge LLMs: GLM-4, Mistral-NeMo and LLaMA3.1, which are applied to text summarization tasks in financial statement analysis. By comparing these models across multiple criteria, we aim to identify strengths, limitations, and use-case suitability for each model. Figure~\ref{fig-PE} illustrates the flow chart of prompt engineering in financial report summarization. The process consists of seven steps, and we typically focus on gathering information and analyzing output. For the former part, we open the collected dataset from TradingView and MSN. For the latter part, we offer an evaluation framework that integrates quantitative metrics with qualitative analysis, contributing to the growing need for systematic LLM evaluation tailored to domain-specific requirements.

In this paper, we also provide insights for future model development, emphasizing the role of prompt engineering and domain adaptation for achieving practical performance levels suitable for real-world financial applications. Thanks to the usage of Ollama\footnote{\url{https://github.com/ollama/ollama}}, the LLMs include Gemini-1.5-pro, Mixtral-8x22b, Claude-3-opus and so on could be added to the backend of our system easily. We list our contributions as follows: 
\begin{enumerate}
    \item We establish a detailed performance benchmark for LLMs in financial reporting, providing a foundation for future comparative studies.
    \item We highlight domain-specific challenges and propose model adjustments to meet the rigorous standards required in finance.
    \item We foster a collaborative environment by openly sharing our dataset and methodologies. Through these efforts, we advance the responsible deployment of LLMs in finance and catalyze ongoing innovation in this area.
\end{enumerate}

\section{Related Works}

\subsection{Large Language Models}

Large Language Models (LLMs) have revolutionized natural language processing, enabling significant advancements in text generation, summarization, translation, and question answering. Models like OpenAI's GPT-3~\cite{gpt-3} and its successor, GPT-4~\cite{gpt-4}, represent breakthroughs in generative modeling due to their ability to generate coherent, contextually relevant text with minimal input. These models have been trained on diverse, large-scale datasets, allowing them to generalize across various domains and tasks. However, their general-purpose nature often requires domain-specific fine-tuning to perform well in specialized areas such as finance, healthcare, and legal domains.

Several research efforts have sought to tailor LLMs to specialized fields. For instance, BioBERT~\cite{biobert} and ClinicalBERT~\cite{clinicalbert} are domain-adapted models for biomedical text, fine-tuned to handle the unique terminology and contextual demands of medical research. Research on smaller yet effective models, such as DistilBERT~\cite{distilbert} and MiniLM~\cite{minilm}, has demonstrated that knowledge distillation and parameter optimization techniques can enhance computational efficiency without significantly compromising performance. These smaller models, along with fine-tuning methods like LoRA~\cite{lora}, support adaptability and efficiency for resource-constrained applications, which is increasingly relevant as LLM applications expand into industries requiring both accuracy and speed.


\subsection{Financial Statement Analysis}

LLM applications in the finance sector are rapidly expanding, with a few pioneering models leading the way. BloombergGPT \cite{BloombergGPT} introduced the first LLM explicitly designed for the finance industry, showcasing the potential for domain-specific language models. To address the need for open-access tools, FinGPT \cite{fingpt} was developed as an open-source, fully featured LLM system. FinGPT provides modules for data collection, model fine-tuning, and cloud-based deployment. PIXIU \cite{pixiu} offers another comprehensive framework for financial analysis, encompassing tasks such as headline classification, question answering, sentiment analysis, and text summarization. Utilizing fine-tuned models based on LLaMA\footnote{\url{https://huggingface.co/TheFinAI/finma-7b-nlp}}, PIXIU is designed to handle complex, multi-faceted tasks within finance. Kim et al.~\cite{finany} explored LLM applications in financial statement analysis, focusing on GPT-4. Their research demonstrated GPT-4’s capacity to analyze standardized and anonymized financial statements, accurately predicting potential future earnings even without narrative or sector-specific context. Remarkably, GPT-4’s performance often surpassed that of traditional financial analysts, reinforcing the potential for LLMs in financial forecasting. 

\subsection{Evaluation of LLMs' Summaries}
In recent years, evaluating LLM performance has relied on established metrics like ROUGE \cite{lin-2004-rouge} and BERTScore~\cite{bertscore}. ROUGE captures $n$-gram overlap, which is useful for content recall and syntactic similarity, while BERTScore leverages contextual embeddings to evaluate semantic similarity, offering a more nuanced assessment of relevance and factual accuracy. 

Previous studies have primarily explored LLM summarization performance in general contexts. For example, Basya et al.\cite{textsum} investigated text summarization with LLMs but focused on general, non-domain-specific text and smaller model architectures, limiting the applicability of findings to high-stakes or specialized fields like finance. More recent research has expanded the scope of LLM evaluation to include domain-specific applications. Afzal et al.~\cite{AdaptEval} evaluated eleven models across medical, scientific, and governmental datasets, prioritizing the domain adaptation capabilities of LLMs. Their findings indicate that smaller LLMs can achieve competitive performance in domain-shift tasks even with minimal training samples, underscoring the potential of compact models in resource-limited or highly specialized settings.



\section{Methodology}\label{Methodology}
\subsection{Preliminaries}
\textbf{Few-Shot Prompting} \cite{fewshot} is a technique in natural language processing where a pre-trained language model is given only a few examples (typically 1 to 5) of a particular task within the input prompt to help the model understand the task it needs to perform. This approach leverages the model's ability to learn from context without requiring fine-tuning on extensive task-specific data. Few-shot prompting has become a popular method for adapting LLMs, especially when there are limited labeled data available for a task.

In few-shot prompting, the language model is presented with a series of example inputs and their corresponding outputs, followed by a new input (the task query) \cite{fewshotpara}. The model then generates an output based on the pattern established in the examples. This method is especially effective with large pre-trained models like GPT-4, which have shown strong generalization capabilities across diverse tasks by only observing a few examples.

\textbf{GLM-4}~\cite{glm2024chatglm} is a prominent LLM that excels across benchmarks in semantics, mathematics, reasoning, coding, and knowledge understanding. As an open-source project under the Apache License, it is suitable for commercial use. GLM-4 utilizes an auto-regressive model to capture local context in input text. Additionally, the end-to-end voice model GLM-4-Voice\footnote{\url{https://github.com/THUDM/GLM-4-Voice/}} was released in October 2024.

\textbf{Mistral-NeMo}\footnote{\url{https://mistral.ai/news/mistral-nemo/}} is a small language model with a 128K context length, making it accessible for third-party developers. Its ``sliding window attention" enhances local comprehension of input text. Recently, Mistral AI has begun transitioning its models to the Mamba architecture~\cite{mamba}. Notably, just one day after the release of LLaMA3.1, Mistral AI unveiled Mistral Large 2, a 123B-parameter model\footnote{\url{https://mistral.ai/news/mistral-large-2407/}} comparable to LLaMA3.1.

\textbf{LLaMA} has significantly advanced the field of LLMs as an open-source model. LLaMA emphasized the critical role of high-quality data in LLM training, influencing many research teams to prioritize diverse data collection. Notably, the 13B-parameter LLaMA outperformed OpenAI’s GPT-3 in multiple benchmarks, underscoring the importance of data over sheer parameter size. The LLaMA series has since expanded rapidly, with LLaMA3.1 reaching 405B parameters. The more recent LLaMA3.2\footnote{\url{https://www.llama.com/}} includes multilingual models (1B, 3B) and text-image models (11B, 90B), with quantized versions for efficiency. Our study utilizes LLaMA3.1 with 70B parameters and 4-bit quantization.    

\subsection{Usage of Prompt Engineering}
Prompt engineering is a commonly utilized technique for extending the capabilities of large language models (LLMs) and has become a comprehensive research topic \cite{sahoo2024systematicsurveypromptengineering} with the development of LLMs. Without updating the core model parameters, this approach leverages task-specific instructions, known as prompts, to enhance model efficacy. Usually, specified prompts are provided together with input to get better results. 

We adopt prompt engineering when we summarize financial reports here and compare the performance among the models we introduced in Section~\ref{Methodology}. Although prompt engineering is commonly utilized in LLM applications, there are no official standards or rules to obey. The beginning of prompts we used to generate the financial reports summarization are shown below: 

\begin{verbatim}
You will be provided with financial notes,
and your task is to summarize the points 
as follows:
\end{verbatim}

In this paper, we use few-shot prompting \cite{fewshot} to make an effective prompt. We specify \textbf{Iruca-GLM-4}, \textbf{Iruca-Mistral-NeNo}, and \textbf{Iruca-LLaMa3.1} in this paper to the models when we utilize prompt engineering with the base models.

\begin{table}[tbp]\scriptsize
\caption{Samples of the Open Dataset}
\label{tab:opendataset}
\centering
\begin{tabular}{m{0.5cm} m{2.cm} m{3.cm} m{1cm}}
    \toprule
    \textbf{Lang}  &\makecell[c]{\textbf{Title}} &\makecell[c]{\textbf{Content}} & \makecell[c]{\textbf{Symbol}}   \\ \midrule
    EN & Will Bitcoin Break September's Jinx? What Data Suggests & Bitcoin, the largest cryptocurrency by market capitalization, ended the month of August down 8.73\%, ...& BTCUSDT \\
    \midrule
    CN & \begin{CJK}{UTF8}{gbsn}【快讯】港股区块链概念股走强，美图涨超6\%\end{CJK} &\begin{CJK}{UTF8}{gbsn}比特币再创历史新高，带动港股市场区块链概念股集体走强。...\end{CJK} & 1357 \\
    \midrule
    JA	& \begin{CJK}{UTF8}{min}東京為替：ドル・円は144円台後半で推移、やや下げ渋る状態が続く\end{CJK} & \begin{CJK}{UTF8}{min}30日午前の東京市場でドル・円は144円80銭近辺で推移。...\end{CJK} & USDJPY   \\
    \bottomrule
\end{tabular}
\end{table}

\subsection{Data Overview}
We source financial data from TradingView and MSN to support our LLMs performance comparison efforts, ensuring a diverse and high-quality dataset for effective model training.

We collect articles covering 10 different financial products, accessing this content through publicly available links on both platforms.\footnote {Copyright information indicates that the dataset's content originates from 55 distinct providers. All rights are retained by the original providers, and this dataset is intended strictly for research purposes, excluding any commercial use.} The samples of our open dataset are shown in Table~\ref{tab:opendataset}, which comprises of 1,000 news articles, providing English (EN), Chinese (CN) and Japanese (JA) languages content (denoted as ``lang''). The ``title'' serves as a brief and clear description of subject matter of the article, the `content' refers to the main body of the article. And ``symbol'' typically denotes unique identifier used to represent a specific commodity in financial market.
Each article contains 17 fields, we primarily utilize the \textbf{``symbol''} and \textbf{``content''} fields to focus on content summarization and domain-specific language adaptation. This dataset serves as a foundational resource for enhancing the contextual relevance and performance of our LLMs in financial reporting tasks.

\subsection{Data Source Selection}

\textbf{Why TradingView and MSN?} TradingView and MSN are established financial aggregation platforms that consolidate real-time information from diverse and reputable financial media sources. Their robust infrastructures enable high-frequency updates, which allows for a continuous stream of relevant data reflective of real market conditions. These platforms also offer public access, making them suitable and accessible sources for gathering a wide range of financial news. By selecting these sources, we ensure our dataset captures a representative snapshot of ongoing market dynamics and sentiment across various asset classes.

\textbf{Why these specific symbols?} We target financial products that are both widely recognized by general users and currently experiencing heightened market attention. Our selection encompasses a diverse mix: market indices that reflect broader economic trends, popular stocks from the U.S. and Japanese markets that attract global interest, widely traded ETFs, major cryptocurrencies, and key forex pairs. This strategic choice ensures that our dataset includes products that are not only prominent but also exhibit varying degrees of volatility, liquidity, and market behavior. Therefore, our strategy can ultimately enhance model’s robustness and relevance in financial applications.


\section{Evaluation and Analysis}

We first clear the effectiveness of our proposed summarization of one practical sample in Table~\ref{tab:sample}. Compared with the raw text, the summary generated becomes obviously simplified, which would be helpful for people such the finance analyst and investigators to reduce the scanning the related news time.

To assess the results of the summarization, we focus on three primary dimensions: \emph{accuracy}, \emph{informativeness} and \emph{coherence}.

\textbf{\emph{Accuracy}} evaluates whether the generated summary faithfully captures essential information and factual content from the original text. To assess the accuracy, we consider the following criteria:

\begin{table}[tbp]
\caption{Summarizaion Sample}
\label{tab:sample}
\centering
\begin{tabular}{m{1cm}|m{6.7cm}}
    \toprule
    \textbf{Type}  &\makecell[c]{\textbf{Content}}   \\ \midrule
    Raw	Text&  RNS Number : 4715Y Toyota Motor Fin (Netherlands) B.V. 30 July 2024 Publication of Annual Financial Report - Toyota Motor Finance (Netherlands) B.V. Toyota Motor Finance (Netherlands) B.V. has published its Annual Financial Report for the financial year ended 31 March 2024. ... \textcolor{cyan}{(1413 Words)}  \\ 
    \midrule
    Summary	&  Toyota Motor Finance (Netherlands) B.V. has published its Annual Financial Report for the financial year ended 31 March 2024. The report can be accessed through the provided link or will be available shortly at the National Storage Mechanism. This information is provided by RNS, the news service of the London Stock Exchange, and is subject to their terms and conditions. \textcolor{cyan}{(373 Words)} \\
    \bottomrule
\end{tabular}
\end{table}

\begin{itemize} \item Are the facts and statements within the summary consistent with those presented in the original text? \item Does the summary emphasize the core points from the original text, especially those central to the topic? \item Importantly, the summary should avoid introducing information not present in the original text, ensuring that readers are not misled by extraneous or fabricated content. \end{itemize}

\textbf{\emph{Informativeness}} measures the extent of useful information included in the summary, i.e., whether the generated summary covers all the important information from the original text. To assess the informativeness, we consider the following criteria:
	
\begin{itemize} 
    \item Does the summary encompass all major points and key details from the original text? 
    \item While simplifying content, does the summary retain sufficient depth and scope to provide a comprehensive understanding? 
    \item It should aim to capture as much relevant information as possible while minimizing redundancy. 
\end{itemize}

\textbf{\emph{Coherence}} examines the logical consistency and fluidity of the information flow within the summary. To assess the coherence, we consider the following criteria:

\begin{itemize} 
    \item Is the content organized in a logical sequence, allowing readers to understand it with ease? 
    \item Does the summary exhibit a smooth, natural flow of sentences, free from abrupt or disjointed transitions? 
    \item It is essential that the summary maintains a consistent theme and narrative style, avoiding fragmented or scattered content. 
\end{itemize}

These three dimensions collectively enable a holistic evaluation of the summarization quality, guiding improvements in LLM performance for generating summaries that are accurate, coherent, and informative.

\subsection{Evaluation Metrics}
Building on our considerations on \emph{accuracy}, \emph{informativeness} and \emph{coherence}, we apply specific evaluation methods to quantitatively and qualitatively assess the quality of generated summaries:

\begin{itemize}
    \item \textbf{Automatic Evaluation}: We utilize established metrics such as ROUGE~\cite{lin-2004-rouge} and BERTScore~\cite{bertscore} to calculate objective scores, providing insights into the generated summaries' quality in terms of similarity to reference texts. ROUGE measures $n$-gram overlap, which is indicative of content recall, while BERTScore leverages contextual embeddings to evaluate semantic similarity, offering a deeper assessment of relevance and accuracy.
    \item \textbf{LLM Evaluation}: We specifically employ Nistral-Nemo, as evaluators for simulating financial experts. By prompting the model with three distinct, manually crafted prompts, we aim to simulate expert-level comparison between the generated summaries and original texts, focusing on qualitative aspects such as depth of understanding, factual alignment, and overall relevance in a financial context.
    
\end{itemize}
	
\begin{table}[tbp]
\caption{Experimental Results on the Three LLMs with Prompt Engineering.}
\label{tab:res}
\centering
\begin{tabular}{l|cccc}
\toprule
Model &\makecell[c]{Iruca-\\GLM-4} &\makecell[c]{Iruca-\\Mistral-NeMo} &\makecell[c]{Iruca-\\LLaMA3.1} \\ 
\midrule
\multicolumn{4}{c}{Evaluation on \emph{Accuracy}} \\
\midrule
ROUGE-1 Precision     & 0.57       & 0.41       & \tb{0.65} \\
BERT Precision        & 0.89       & 0.85       & \tb{0.91} \\
LLM Score Precision   & \tb{0.80}  & 0.79       & 0.73 \\
Iteration Time (s)    & 8.25       & \tb{3.02}  & 63.35 \\
\midrule
\multicolumn{4}{c}{Evaluation on \emph{Informativeness}} \\
\midrule
ROUGE-1 Recall      & 0.50      & 0.30       & \tb{0.51}  \\
BERT Recall         & \tb{0.88} & 0.85       & \tb{0.88}  \\
LLM Score Recall    & 0.80      & 0.80       & \tb{0.83}  \\
Iteration Time (s)  & 8.16      & \tb{2.99}  & 23.72 \\
\midrule
\multicolumn{4}{c}{Evaluation on \emph{Coherence}} \\
\midrule
ROUGE-1 F1          & 0.47       & 0.31       & \tb{0.52}  \\
BERT F1             & \tb{0.89}  & 0.85       & \tb{0.89}  \\
LLM Score F1        & 0.81       & 0.78       & \tb{0.83}  \\
Iteration Time (s)  & 8.07       & \tb{3.04}  & 25.04 \\
\bottomrule
\end{tabular}
\end{table}


\textbf{Measurement of \emph{Accuracy}}. Precision quantifies how many of the elements in the generated output are relevant and correct with respect to the reference text. Here, we use ROUGE-1 Precision and BERT Precision. Although ROUGE-1 Precision and LLM Score Precision share a indirect relationship, higher ROUGE-1 Precision typically contributes to an increased BERT Precision. Furthermore, a strong positive correlation exists between BERT Precision and LLM Score Precision, indicating that enhanced semantic matching frequently leads to improved overall summary quality.

\textbf{Measurement of \emph{Informativeness}}. Recall measures how much relevant content from the reference (original) text is captured in the generated output. Recall measures informativeness in LLM-based summarization by ensuring that a summary captures most of the relevant details and ideas from the source text. High recall indicates comprehensive coverage, a critical component of informativeness.

\textbf{Measurement of \emph{Coherence}}. F1 score is the harmonic mean of precision and recall, integrating both metrics into a single measure of model performance. The F1 score in summarization tasks for LLMs helps assess coherence by balancing the precision (relevance) and recall (coverage) of critical information in the generated summary, providing a straightforward but effective metric to evaluate how well the summary aligns with the source content.

\subsection{Experimental Results}
The evaluation results for our models demonstrate several key insights regarding their performance in text summarization tasks.

As shown in Table~\ref{tab:res} for evaluation on \emph{accuracy}, in terms of ROUGE-1 Precision, Iruca-LLaMA3.1 achieves the highest score (0.65), indicating strong token-level alignment with reference texts, followed by Iruca-GLM-4 (0.57) and Iruca-Mistral-NeMo (0.41). For BERT Precision, Iruca-LLaMA3.1 again leads with a score of 0.91, closely followed by Iruca-GLM-4 at 0.89, and Iruca-Mistral-NeMo at 0.85, highlighting strong semantic accuracy across all models. The LLM Score Precision shows slightly different results, with Iruca-GLM-4 achieving the highest score (0.80), marginally outperforming Iruca-Mistral-NeMo (0.79) and Iruca-LLaMA3.1 (0.73). This suggests that Iruca-GLM-4 may offer better holistic quality as judged by this specific metric. Regarding iteration time, Iruca-LLaMA3.1 is significantly slower (63.35s) compared to Iruca-GLM-4 (8.25s) and Iruca-Mistral-NeMo (3.02s), indicating a trade-off between accuracy and computational efficiency.

In terms of ROUGE-1 Recall shown in Table~\ref{tab:res} for evaluation on \emph{informativeness}, both Iruca-LLaMA3.1 and Iruca-GLM-4 achieve high scores (0.51 and 0.50, respectively), coupled with strong BERT Recall values (0.88), indicating comprehensive vocabulary coverage and robust semantic alignment. These results suggest that elevated ROUGE-1 Recall often correlates with higher BERT Recall, implying that these models not only capture a broad range of vocabulary in their summaries but also maintain substantial semantic congruence with reference texts. Interestingly, the LLM Score Recall remains comparable across all models, suggesting that this metric reflects overall summary quality rather than simple vocabulary overlap. Therefore, while models with higher ROUGE-1 Recall (such as Iruca-LLaMA3.1 and Iruca-GLM-4) exhibit broader coverage, this does not necessarily result in a higher LLM Score Recall, as the latter encompasses a more holistic quality assessment.

In Table~\ref{tab:res} for evaluation on \emph{coherence}, the ROUGE-L F1 Score for Iruca-LLaMA3.1 is 0.52, with a BERT F1 Score of 0.89, demonstrating strong coherence, sequential accuracy, and effective semantic matching. This relationship suggests that higher coherence and sequence alignment often correlate with improved semantic accuracy. In contrast, Iruca-GLM-4 achieves a slightly lower ROUGE-L F1 Score of 0.47, while maintaining a high BERT F1 Score of 0.89. This indicates that, although the model’s sequential alignment may be diminished, it retains semantic accuracy, effectively capturing the meaning of reference summaries. Iruca-Mistral-NeMo, however, presents a ROUGE-L F1 Score of 0.31 and a BERT F1 Score of 0.85, suggesting that its lower sequence alignment may adversely impact its semantic matching capabilities.

Across models, we observe a positive correlation between ROUGE-L F1 and LLM F1 Scores, suggesting that higher coherence aligns with improved overall quality ratings. Similarly, the strong correlation between BERT F1 and LLM F1 Scores indicates that better semantic alignment leads to higher quality evaluations.

In summary, our findings reveal that both ROUGE and BERT scores are positively correlated, capturing vocabulary coverage and semantic matching. Coherence and sequence information, as indicated by ROUGE-L F1, contribute significantly to semantic alignment and overall quality. The LLM Scores, as comprehensive metrics, offer valuable insights into the factors impacting summarization performance, underscoring their importance as benchmarks for model evaluation.

	

\section{Conclusion and Future Works}
This paper presents an empirical evaluation of GLM-4, Mistral-NeMo and LLaMA3.1 for text summarization tasks in the financial domain. Through a comprehensive analysis, we have identified the strengths and trade-offs associated with each model in terms of accuracy, coherence, and informativeness.

Our findings indicate the following for summarizing financial reports:
\begin{itemize}
    \item \textbf{Iruca-GLM-4} strikes a balance between speed and accuracy, Iruca-GLM-4 proves to be a versatile option. It is well-suited for tasks that require high-quality text within a moderate time frame, making it a practical option for most financial summarization needs.
    
    \item \textbf{Iruca-Mistral-NeMo} is with a significant advantage in speed, this model offers the fastest summarization but may sacrifice some text quality in specific dimensions. If quick response time is the primary criterion, Iruca-Mistral-NeMo serves as an efficient choice.
    
    \item \textbf{Iruca-LLaMA3.1} demonstrates exceptional accuracy and high-quality text generation, albeit with the longest processing time. For tasks where accuracy and nuanced text quality are paramount and time constraints are minimal, Iruca-LLaMA3.1 is the preferred choice.
\end{itemize}

In our ongoing efforts to enhance LLM performance, we are actively engaged in fine-tuning to meet the practical requirements of our business partners, aiming for more accurate, relevant, and efficient outputs in real-world financial applications. Therefore, for the future work, we plan to explore knowledge distillation techniques to develop a finance-specific Small Language Model (SML), which would enable streamlined, domain-focused capabilities with lower computational demands.

Besides, we notice that explainability remains a crucial and enduring challenge in LLM research, especially for applications in high-stakes fields like finance where trust and interpretability are paramount. We are committed to contributing to explainability research, focusing on methods that provide clearer insights into model decisions and align generated outputs with financial reasoning.

Lastly, to further validate and refine our models, we will collaborate with financial experts who will review and compare the generated summaries against original texts. Their expert feedback will be invaluable in assessing content accuracy, contextual appropriateness, and overall value, ultimately guiding us toward more robust and reliable LLMs in financial summarization tasks.

\bibliographystyle{IEEEtran}
\bibliography{refs}	

\appendix
\textbf{Data Sources and Content Providers.} The specific data sources used in our study are as follows:
\begin{enumerate}
    \item \url{https://jp.tradingview.com/news/}.
    \item \url{https://cn.tradingview.com/news/}.
    \item \url{https://www.tradingview.com/news/}.
    \item \url{https://www.msn.com/ja-jp/money/}.
    \item \url{https://www.msn.com/ja-jp/news/}.
\end{enumerate}

The 55 content providers are given as follows:

``Kagoshima Television Broadcasting'', ``dow-jones'', ``Finasee'', ``Financial Field'', ``San-in Chuo Television'', ``JBpress'', ``forexlive'', ``Yomiuri Shimbun'', ``gelonghui'', ``KOREA WAVE'', ``cryptopotato'', ``u today'', ``THE GOLD ONLINE'', ``Bloomberg'', ``Toyama Television'', ``Jiji Press'', ``gurufocus'', ``moneycontrol'', ``benzinga'', ``mt-newswires'', ``TV Asahi news'', ``zacks'', ``Television Nishinippon'', ``cointelegraph'', ``cryptonews'', ``tradingview'', ``trading-economics'', ``Asahi Shimbun Digital'', ``NNA Asia Economic News'', ``FNN Prime Online'', ``Hokkaido Broadcasting'', ``barchart'', ``coindesk'', ``Kyodo News'', ``market-watch'', ``Tokyo Shimbun'', ``fisco'', ``todayq'', ``Nippon Television NEWS NNN'', ``TBS NEWS'', ``Ibaraki Shimbun Cross Eye'', ``newsbtc'', ``Toyo Keizai Online'', ``cryptoglobe'', ``Reuters'', ``macenews'', ``reuters'', ``Sankei Shimbun'', ``beincrypto'', ``the block'', ``coindeskjapan'', ``KBC Kyushu Asahi Broadcasting'', ``All About'', ``Mainichi Shimbun'', ``fisco''.

\end{document}